\RequirePackage{snapshot}

\documentclass[runningheads]{llncs}
\usepackage{subcaption}
\usepackage{graphicx}

\usepackage{tikz}
\usepackage{comment}
\usepackage{amsmath,amssymb} %
\usepackage{color}
\usepackage{booktabs}

\usepackage{xfrac}
\usepackage{soul}
\usepackage{placeins}
\usepackage{wrapfig}
\usepackage{multirow}
\usepackage[export]{adjustbox}
\usepackage{pifont}

\newcommand{\diffusionmodel}{\emph{RDM}\,}
\newcommand{\cf}{\emph{cf.}\,}

\providecommand{\imwidth}{}
\providecommand{\impath}[1]{}
\providecommand{\impatha}[1]{}
\providecommand{\impathb}[1]{}
\providecommand{\impathc}[1]{}
\providecommand{\impathd}[1]{}

\newcommand{\neatsamples}{
\begin{figure}[htbp]
\renewcommand{\imwidth}{.31\textwidth}
\renewcommand{\impath}[1]{img/goodsamples/##1}
\centering
\begin{tabular}{c@{\hspace{0pt}}c@{\hspace{0pt}}c}
\toprule
\includegraphics[width=\imwidth]{\impath{the_holy_grail_filled_with_the_essence_of_consciousness}} &
\includegraphics[width=\imwidth]{\impath{the_rise_of_consciousness}} &
\includegraphics[width=\imwidth]{\impath{a_red_sun_is_drowning}} \\

\shortstack{\tiny \emph{'The Holy Grail filled with} \\ \tiny \emph{the essence of consciousness.'}} &
\tiny \emph{'The rise of consciousness.'} &
\tiny \emph{'A red sun is drowning.'} \\

\includegraphics[width=\imwidth]{\impath{the_central_nervous_system}} &
\includegraphics[width=\imwidth]{\impath{00418}} &
\includegraphics[width=\imwidth]{\impath{humanity_is_killed_by_artificial_intelligence}} \\

\tiny \emph{'The central nervous system.'} &
\shortstack{\tiny \emph{'Making AI generated images releases} \\ %
\tiny \emph{endorphins, a children's story book.'}} &
\shortstack{\tiny \emph{'Humanity is killed} \\ \tiny \emph{by artificial intelligence.'}} \\

\includegraphics[width=\imwidth]{\impath{00306}} &
\includegraphics[width=\imwidth]{\impath{fear_of_the_dark}} &
\includegraphics[width=\imwidth]{\impath{00544}} \\

\shortstack{\tiny \emph{'Stones of eyes in color} \\ \tiny \emph{of hue's of hunts and colors.'}} &
\tiny \emph{'Fear of the dark.'} &
\tiny \emph{'Magic tree of life.'} \\
\bottomrule

\end{tabular}

\caption{\label{fig:bigsamples} Selected $768 \times 768$ samples for various text inputs obtained with our ``stylized'' LAION-RDM  (see Sec.~\ref{subsec:stylization}). Note that the model was only trained on images.%
}
\end{figure}

}

\newcommand{\neatersamples}{
\begin{figure}[htbp]
\renewcommand{\imwidth}{.31\textwidth}
\renewcommand{\impath}[1]{img/goodsamples2/##1}
\centering
\begin{tabular}{c@{\hspace{0pt}}c@{\hspace{0pt}}c}
\toprule
\includegraphics[width=\imwidth]{\impath{a_fleet_in_the_eye_of_the_storm}} &
\includegraphics[width=\imwidth]{\impath{a_star_is_exploding}} &
\includegraphics[width=\imwidth]{\impath{fire_and_ice_collide2}} \\

\shortstack{\tiny \emph{'A fleet in} \\ \tiny \emph{the eye of the storm.'}} &
\tiny \emph{'A star is exploding.'} &
\tiny \emph{'Fire and ice collide.'} \\

\includegraphics[width=\imwidth]{\impath{incarnation_of_the_outrages_of_man}} &
\includegraphics[width=\imwidth]{\impath{lion_king_of_beasts}} &
\includegraphics[width=\imwidth]{\impath{the_temple_pf_twilight}} \\

\tiny \emph{'Incarnation of the outrages of men.'} &
\tiny \emph{'Lion, king of beasts.} & %
\tiny \emph{'The temple of twilight.} \\

\includegraphics[width=\imwidth]{\impath{twilight_of_the_gods}} &
\includegraphics[width=\imwidth]{\impath{00070}} &
\includegraphics[width=\imwidth]{\impath{00055}} \\

\tiny \emph{'Twilight of the Gods.'} &
\tiny \emph{'Warp speed.'} &
\shortstack{\tiny \emph{'Stones of eyes in color of} \\ \tiny \emph{hue's of hunts and colors.'}}  \\
\bottomrule

\end{tabular}

\caption{\label{fig:bigsamples2} More $768 \times 768$ samples for various text inputs obtained with our ``stylized'' LAION-RDM  (see Sec.~\ref{subsec:stylization}). Note that the model was only trained on images.%
}
\end{figure}

}

\newcommand{\finegrainedsupp}{
\renewcommand{\imwidth}{.08\textwidth}
\renewcommand{\impatha}[1]{img/stylization/holy_grail/##1}
\renewcommand{\impathc}[1]{img/stylization/creation_of_world/##1}
\renewcommand{\impathb}[1]{img/stylization/conquest_of_paradise/##1}
\renewcommand{\impathd}[1]{img/stylization/twilight_of_the_gods/##1}
\begin{figure}[htbp]
\setlength{\tabcolsep}{1pt}
\begin{tabular}{c@{\hspace{2pt}}cccccccccc}
\toprule
&\shortstack{\tiny \emph{art} \\ \tiny \emph{nouveau}} & \parbox[c]{.1\textwidth}{\centering \vspace{-.6em} \tiny \emph{baroque} }& \shortstack{\tiny \emph{expressio-} \\ \tiny \emph{nism}} & \shortstack{\tiny \emph{impres-} \\ \tiny \emph{sionism}}& \shortstack{\tiny \emph{post-imp-}  \\ \tiny \emph{ressionism}} &
\parbox[c]{.1\textwidth}{\centering \vspace{-.6em} \tiny \emph{realism}}& \shortstack{ \tiny \emph{renais-} \\ \tiny \emph{sance}}  & \shortstack{ \tiny \emph{roman-} \\ \tiny \emph{ticism}} & \shortstack{\tiny \emph{sur-} \\ \tiny \emph{realism}} & \parbox[c]{.1\textwidth}{\centering \vspace{-.6em} \tiny \emph{ukiyo-e}} \\ 
\midrule
\multirow{2}{*}[0.045\textwidth]{\rotatebox[origin=c]{90}{\shortstack{\tiny \emph{'The holy grail filled} \\ \tiny \emph{with the essence} \\ \tiny \emph{of consciousness.'}}}} &
\includegraphics[width=\imwidth]{\impatha{batched_samples-style_art_nouveau-run0-sample2-prompt_The_holy_grail_filled_with_the_essence_of_consciousness}} &
\includegraphics[width=\imwidth]{\impatha{batched_samples-style_baroque-run0-sample0-prompt_The_holy_grail_filled_with_the_essence_of_consciousness}} &
\includegraphics[width=\imwidth]{\impatha{batched_samples-style_expressionism-run0-sample2-prompt_The_holy_grail_filled_with_the_essence_of_consciousness}} &
\includegraphics[width=\imwidth]{\impatha{batched_samples-style_impressionism-run0-sample0-prompt_The_holy_grail_filled_with_the_essence_of_consciousness}} &
\includegraphics[width=\imwidth]{\impatha{batched_samples-style_post_impressionism-run0-sample0-prompt_The_holy_grail_filled_with_the_essence_of_consciousness}} &
\includegraphics[width=\imwidth]{\impatha{batched_samples-style_realism-run0-sample0-prompt_The_holy_grail_filled_with_the_essence_of_consciousness}} &
\includegraphics[width=\imwidth]{\impatha{batched_samples-style_renaissance-run0-sample2-prompt_The_holy_grail_filled_with_the_essence_of_consciousness}} &
\includegraphics[width=\imwidth]{\impatha{batched_samples-style_romanticism-run0-sample1-prompt_The_holy_grail_filled_with_the_essence_of_consciousness}} &
\includegraphics[width=\imwidth]{\impatha{batched_samples-style_surrealism-run0-sample0-prompt_The_holy_grail_filled_with_the_essence_of_consciousness}} &
\includegraphics[width=\imwidth]{\impatha{batched_samples-style_ukiyo_e-run0-sample2-prompt_The_holy_grail_filled_with_the_essence_of_consciousness}} \\[-3pt]

&
\includegraphics[width=\imwidth]{\impatha{batched_samples-style_art_nouveau-run0-sample3-prompt_The_holy_grail_filled_with_the_essence_of_consciousness}} &
\includegraphics[width=\imwidth]{\impatha{batched_samples-style_baroque-run0-sample2-prompt_The_holy_grail_filled_with_the_essence_of_consciousness}} &
\includegraphics[width=\imwidth]{\impatha{batched_samples-style_expressionism-run0-sample3-prompt_The_holy_grail_filled_with_the_essence_of_consciousness}} &
\includegraphics[width=\imwidth]{\impatha{batched_samples-style_impressionism-run0-sample3-prompt_The_holy_grail_filled_with_the_essence_of_consciousness}} &
\includegraphics[width=\imwidth]{\impatha{batched_samples-style_post_impressionism-run0-sample2-prompt_The_holy_grail_filled_with_the_essence_of_consciousness}} &
\includegraphics[width=\imwidth]{\impatha{batched_samples-style_realism-run0-sample2-prompt_The_holy_grail_filled_with_the_essence_of_consciousness}} &
\includegraphics[width=\imwidth]{\impatha{batched_samples-style_renaissance-run0-sample3-prompt_The_holy_grail_filled_with_the_essence_of_consciousness}} &
\includegraphics[width=\imwidth]{\impatha{batched_samples-style_romanticism-run0-sample2-prompt_The_holy_grail_filled_with_the_essence_of_consciousness}} &
\includegraphics[width=\imwidth]{\impatha{batched_samples-style_surrealism-run0-sample3-prompt_The_holy_grail_filled_with_the_essence_of_consciousness}} &
\includegraphics[width=\imwidth]{\impatha{batched_samples-style_ukiyo_e-run0-sample3-prompt_The_holy_grail_filled_with_the_essence_of_consciousness}} \\
\midrule

\multirow{2}{*}[0.025\textwidth]{\rotatebox[origin=c]{90}{\shortstack{\tiny \emph{'The conquest} \\ \tiny \emph{of paradise.'}}}} &
\includegraphics[width=\imwidth]{\impathb{batched_samples-style_art_nouveau-run0-sample0-prompt_The_conquest_of_paradise}} &
\includegraphics[width=\imwidth]{\impathb{batched_samples-style_baroque-run0-sample0-prompt_The_conquest_of_paradise}} &
\includegraphics[width=\imwidth]{\impathb{batched_samples-style_expressionism-run0-sample0-prompt_The_conquest_of_paradise}} &
\includegraphics[width=\imwidth]{\impathb{batched_samples-style_impressionism-run0-sample2-prompt_The_conquest_of_paradise}} &
\includegraphics[width=\imwidth]{\impathb{batched_samples-style_post_impressionism-run0-sample0-prompt_The_conquest_of_paradise}} &
\includegraphics[width=\imwidth]{\impathb{batched_samples-style_realism-run0-sample0-prompt_The_conquest_of_paradise}} &
\includegraphics[width=\imwidth]{\impathb{batched_samples-style_renaissance-run0-sample0-prompt_The_conquest_of_paradise}} &
\includegraphics[width=\imwidth]{\impathb{batched_samples-style_romanticism-run0-sample0-prompt_The_conquest_of_paradise}} &
\includegraphics[width=\imwidth]{\impathb{batched_samples-style_surrealism-run0-sample2-prompt_The_conquest_of_paradise}} &
\includegraphics[width=\imwidth]{\impathb{batched_samples-style_ukiyo_e-run0-sample0-prompt_The_conquest_of_paradise}} \\[-3pt]

&
\includegraphics[width=\imwidth]{\impathb{batched_samples-style_art_nouveau-run0-sample1-prompt_The_conquest_of_paradise}} &
\includegraphics[width=\imwidth]{\impathb{batched_samples-style_baroque-run0-sample3-prompt_The_conquest_of_paradise}} &
\includegraphics[width=\imwidth]{\impathb{batched_samples-style_expressionism-run0-sample3-prompt_The_conquest_of_paradise}} &
\includegraphics[width=\imwidth]{\impathb{batched_samples-style_impressionism-run0-sample3-prompt_The_conquest_of_paradise}} &
\includegraphics[width=\imwidth]{\impathb{batched_samples-style_post_impressionism-run0-sample1-prompt_The_conquest_of_paradise}} &
\includegraphics[width=\imwidth]{\impathb{batched_samples-style_realism-run0-sample1-prompt_The_conquest_of_paradise}} &
\includegraphics[width=\imwidth]{\impathb{batched_samples-style_renaissance-run0-sample2-prompt_The_conquest_of_paradise}} &
\includegraphics[width=\imwidth]{\impathb{batched_samples-style_romanticism-run0-sample2-prompt_The_conquest_of_paradise}} &
\includegraphics[width=\imwidth]{\impathb{batched_samples-style_surrealism-run0-sample3-prompt_The_conquest_of_paradise}} &
\includegraphics[width=\imwidth]{\impathb{batched_samples-style_ukiyo_e-run0-sample1-prompt_The_conquest_of_paradise}} \\

\midrule

\multirow{2}{*}[0.03\textwidth]{\rotatebox[origin=c]{90}{\shortstack{\tiny \emph{'The first day} \\ \tiny \emph{of the creation} \\ \tiny \emph{of the world.'}}}} &
\includegraphics[width=\imwidth]{\impathc{batched_samples-style_art_nouveau-run0-sample1-prompt_The_first_day_of_the_creation_of_the_world}} &
\includegraphics[width=\imwidth]{\impathc{batched_samples-style_baroque-run0-sample0-prompt_The_first_day_of_the_creation_of_the_world}} &
\includegraphics[width=\imwidth]{\impathc{batched_samples-style_expressionism-run0-sample0-prompt_The_first_day_of_the_creation_of_the_world}} &
\includegraphics[width=\imwidth]{\impathc{batched_samples-style_impressionism-run0-sample0-prompt_The_first_day_of_the_creation_of_the_world}} &
\includegraphics[width=\imwidth]{\impathc{batched_samples-style_post_impressionism-run0-sample1-prompt_The_first_day_of_the_creation_of_the_world}} &
\includegraphics[width=\imwidth]{\impathc{batched_samples-style_realism-run0-sample1-prompt_The_first_day_of_the_creation_of_the_world}} &
\includegraphics[width=\imwidth]{\impathc{batched_samples-style_renaissance-run0-sample0-prompt_The_first_day_of_the_creation_of_the_world}} &
\includegraphics[width=\imwidth]{\impathc{batched_samples-style_romanticism-run0-sample0-prompt_The_first_day_of_the_creation_of_the_world}} &
\includegraphics[width=\imwidth]{\impathc{batched_samples-style_surrealism-run0-sample0-prompt_The_first_day_of_the_creation_of_the_world}} &
\includegraphics[width=\imwidth]{\impathc{batched_samples-style_ukiyo_e-run0-sample1-prompt_The_first_day_of_the_creation_of_the_world}} \\[-3pt]

&
\includegraphics[width=\imwidth]{\impathc{batched_samples-style_art_nouveau-run0-sample3-prompt_The_first_day_of_the_creation_of_the_world}} &
\includegraphics[width=\imwidth]{\impathc{batched_samples-style_baroque-run0-sample1-prompt_The_first_day_of_the_creation_of_the_world}} &
\includegraphics[width=\imwidth]{\impathc{batched_samples-style_expressionism-run0-sample2-prompt_The_first_day_of_the_creation_of_the_world}} &
\includegraphics[width=\imwidth]{\impathc{batched_samples-style_impressionism-run0-sample2-prompt_The_first_day_of_the_creation_of_the_world}} &
\includegraphics[width=\imwidth]{\impathc{batched_samples-style_post_impressionism-run0-sample3-prompt_The_first_day_of_the_creation_of_the_world}} &
\includegraphics[width=\imwidth]{\impathc{batched_samples-style_realism-run0-sample3-prompt_The_first_day_of_the_creation_of_the_world}} &
\includegraphics[width=\imwidth]{\impathc{batched_samples-style_renaissance-run0-sample1-prompt_The_first_day_of_the_creation_of_the_world}} &
\includegraphics[width=\imwidth]{\impathc{batched_samples-style_romanticism-run0-sample3-prompt_The_first_day_of_the_creation_of_the_world}} &
\includegraphics[width=\imwidth]{\impathc{batched_samples-style_surrealism-run0-sample2-prompt_The_first_day_of_the_creation_of_the_world}} &
\includegraphics[width=\imwidth]{\impathc{batched_samples-style_ukiyo_e-run0-sample3-prompt_The_first_day_of_the_creation_of_the_world}} \\

\midrule

\multirow{2}{*}[0.025\textwidth]{\rotatebox[origin=c]{90}{\shortstack{\tiny \emph{'Twilight of} \\ \tiny \emph{the gods.'}}}} &
\includegraphics[width=\imwidth]{\impathd{batched_samples-style_art_nouveau-run0-sample2-prompt_Twilight_of_the_Gods}} &
\includegraphics[width=\imwidth]{\impathd{batched_samples-style_baroque-run0-sample1-prompt_Twilight_of_the_Gods}} &
\includegraphics[width=\imwidth]{\impathd{batched_samples-style_expressionism-run0-sample1-prompt_Twilight_of_the_Gods}} &
\includegraphics[width=\imwidth]{\impathd{batched_samples-style_impressionism-run0-sample0-prompt_Twilight_of_the_Gods}} &
\includegraphics[width=\imwidth]{\impathd{batched_samples-style_post_impressionism-run0-sample1-prompt_Twilight_of_the_Gods}} &
\includegraphics[width=\imwidth]{\impathd{batched_samples-style_realism-run0-sample1-prompt_Twilight_of_the_Gods}} &
\includegraphics[width=\imwidth]{\impathd{batched_samples-style_renaissance-run0-sample0-prompt_Twilight_of_the_Gods}} &
\includegraphics[width=\imwidth]{\impathd{batched_samples-style_romanticism-run0-sample1-prompt_Twilight_of_the_Gods}} &
\includegraphics[width=\imwidth]{\impathd{batched_samples-style_surrealism-run0-sample1-prompt_Twilight_of_the_Gods}} &
\includegraphics[width=\imwidth]{\impathd{batched_samples-style_ukiyo_e-run0-sample0-prompt_Twilight_of_the_Gods}} \\[-3pt]

&
\includegraphics[width=\imwidth]{\impathd{batched_samples-style_art_nouveau-run0-sample3-prompt_Twilight_of_the_Gods}} &
\includegraphics[width=\imwidth]{\impathd{batched_samples-style_baroque-run0-sample3-prompt_Twilight_of_the_Gods}} &
\includegraphics[width=\imwidth]{\impathd{batched_samples-style_expressionism-run0-sample2-prompt_Twilight_of_the_Gods}} &
\includegraphics[width=\imwidth]{\impathd{batched_samples-style_impressionism-run0-sample3-prompt_Twilight_of_the_Gods}} &
\includegraphics[width=\imwidth]{\impathd{batched_samples-style_post_impressionism-run0-sample3-prompt_Twilight_of_the_Gods}} &
\includegraphics[width=\imwidth]{\impathd{batched_samples-style_realism-run0-sample2-prompt_Twilight_of_the_Gods}} &
\includegraphics[width=\imwidth]{\impathd{batched_samples-style_renaissance-run0-sample3-prompt_Twilight_of_the_Gods}} &
\includegraphics[width=\imwidth]{\impathd{batched_samples-style_romanticism-run0-sample3-prompt_Twilight_of_the_Gods}} &
\includegraphics[width=\imwidth]{\impathd{batched_samples-style_surrealism-run0-sample2-prompt_Twilight_of_the_Gods}} &
\includegraphics[width=\imwidth]{\impathd{batched_samples-style_ukiyo_e-run0-sample1-prompt_Twilight_of_the_Gods}} \\
\bottomrule
\end{tabular}
\vspace{-1em}
\caption{\label{fig:finegrained_vis_supp} More visual examples of stylization with our LAION model as in Fig.~\ref{fig:finegrained_vis}. \vspace{-2em}}
\end{figure}

}

\newcommand{\stylizer}{
\renewcommand{\imwidth}{.05\textwidth}
\renewcommand{\impatha}[1]{img/stylization/stag/##1}
\renewcommand{\impathb}[1]{img/stylization/fruit_basket/##1}
\renewcommand{\impathc}[1]{img/stylization/woman_piano/##1}
\renewcommand{\impathd}[1]{img/stylization/set_table/##1}
\setlength{\tabcolsep}{1pt}
\begin{wrapfigure}{r}{.5\textwidth}
\vspace{-1em}
\centering
\begin{tabular}{c@{\hspace{2pt}}c@{\hspace{0pt}}c c@{\hspace{0pt}}c c@{\hspace{0pt}}c c@{\hspace{0pt}}c}
\toprule
&\multicolumn{2}{c}{\parbox[c]{.1\textwidth}{\centering \vspace{-.8em} \tiny\emph{'A stag.'}}} & 
\multicolumn{2}{c}{\shortstack{\tiny\emph{'A basket } \\
\tiny\emph{full of fruits.'}}}& 
\multicolumn{2}{c}{\shortstack{\tiny\emph{'A woman } \\
\tiny\emph{playing piano.'}}} & 
\multicolumn{2}{c}{\parbox[c]{.1\textwidth}{\centering \vspace{-.8em} \tiny\emph{'A table set.'}}} \\
\midrule
\multirow{2}{*}[0.01\textwidth]{\rotatebox[origin=c]{90}{$\mathcal{D}_{\text{\tiny style}}$}} &
{\includegraphics[width=\imwidth]{\impatha{style1}}} &
{\includegraphics[width=\imwidth]{\impatha{style2}}} &
{\includegraphics[width=\imwidth]{\impathb{style1}}} &
{\includegraphics[width=\imwidth]{\impathb{style3}}} &
{\includegraphics[width=\imwidth]{\impathc{style1}}} &
{\includegraphics[width=\imwidth]{\impathc{style2}}} &
{\includegraphics[width=\imwidth]{\impathd{style5}}} &
{\includegraphics[width=\imwidth]{\impathd{style1}}} \\ [-3pt]

&
{\includegraphics[width=\imwidth]{\impatha{style5}}} &
{\includegraphics[width=\imwidth]{\impatha{style3}}} &
{\includegraphics[width=\imwidth]{\impathb{style4}}} &
{\includegraphics[width=\imwidth]{\impathb{style5}}} &
{\includegraphics[width=\imwidth]{\impathc{style3}}} &
{\includegraphics[width=\imwidth]{\impathc{style5}}}&
{\includegraphics[width=\imwidth]{\impathd{style2}}} &
{\includegraphics[width=\imwidth]{\impathd{style4}}} \\

\midrule

\multirow{2}{*}[0.01\textwidth]{\rotatebox[origin=c]{90}{$\mathcal{D}_{\text{\tiny train}}$}} &
{\includegraphics[width=\imwidth]{\impatha{orig1}}} &
{\includegraphics[width=\imwidth]{\impatha{orig2}}} &
{\includegraphics[width=\imwidth]{\impathb{orig1}}}&
{\includegraphics[width=\imwidth]{\impathb{orig2}}} &
{\includegraphics[width=\imwidth]{\impathc{orig1}}} &
{\includegraphics[width=\imwidth]{\impathc{orig2}}} &
{\includegraphics[width=\imwidth]{\impathd{orig1}}} &
{\includegraphics[width=\imwidth]{\impathd{orig2}}} \\[-3pt]

&
{\includegraphics[width=\imwidth]{\impatha{orig3}}} &
{\includegraphics[width=\imwidth]{\impatha{orig4}}} &
{\includegraphics[width=\imwidth]{\impathb{orig3}}} &
{\includegraphics[width=\imwidth]{\impathb{orig5}}} &
{\includegraphics[width=\imwidth]{\impathc{orig3}}} &
{\includegraphics[width=\imwidth]{\impathc{orig4}}} &
{\includegraphics[width=\imwidth]{\impathd{orig3}}} &
{\includegraphics[width=\imwidth]{\impathd{orig4}}} \\

\bottomrule
\end{tabular}
\caption{\label{fig:stylize} Zero-shot text-guided stylization with our ImageNet-\diffusionmodel. Best viewed when zoomed in. \vspace{-3em}}
\end{wrapfigure}
}

\newcommand{\finegrainedvisual}{
\vspace{-2em}
\begin{figure}[htbp]
\renewcommand{\imwidth}{.09\textwidth}
\renewcommand{\impath}[1]{img/stylization/day_and_night/##1}
\setlength{\tabcolsep}{1pt}
\begin{tabular}{cccccccccc}
\toprule
\shortstack{\tiny \emph{art} \\ \tiny \emph{nouveau}} & \parbox[c]{.1\textwidth}{\centering \vspace{-.6em} \tiny \emph{baroque} }& \shortstack{\tiny \emph{expressio-} \\ \tiny \emph{nism}} & \shortstack{\tiny \emph{impres-} \\ \tiny \emph{sionism}}& \shortstack{\tiny \emph{post-imp-}  \\ \tiny \emph{ressionism}} &
\parbox[c]{.1\textwidth}{\centering \vspace{-.6em} \tiny \emph{realism}}& \shortstack{ \tiny \emph{renais-} \\ \tiny \emph{sance}}  & \shortstack{ \tiny \emph{roman-} \\ \tiny \emph{ticism}} & \shortstack{\tiny \emph{sur-} \\ \tiny \emph{realism}} & \parbox[c]{.1\textwidth}{\centering \vspace{-.6em} \tiny \emph{ukiyo-e}} \\ 
\midrule
\includegraphics[width=\imwidth]{\impath{batched_samples-style_art_nouveau-run0-sample1-prompt_Day_and_night_fight_for_the_domination_of_time}} &
\includegraphics[width=\imwidth]{\impath{batched_samples-style_baroque-run0-sample1-prompt_Day_and_night_fight_for_the_domination_of_time}} &
\includegraphics[width=\imwidth]{\impath{batched_samples-style_expressionism-run0-sample1-prompt_Day_and_night_fight_for_the_domination_of_time}} &
\includegraphics[width=\imwidth]{\impath{batched_samples-style_impressionism-run0-sample1-prompt_Day_and_night_fight_for_the_domination_of_time}} &
\includegraphics[width=\imwidth]{\impath{batched_samples-style_post_impressionism-run0-sample1-prompt_Day_and_night_fight_for_the_domination_of_time}} &
\includegraphics[width=\imwidth]{\impath{batched_samples-style_realism-run0-sample1-prompt_Day_and_night_fight_for_the_domination_of_time}} &
\includegraphics[width=\imwidth]{\impath{batched_samples-style_renaissance-run0-sample0-prompt_Day_and_night_fight_for_the_domination_of_time}} &
\includegraphics[width=\imwidth]{\impath{batched_samples-style_romanticism-run0-sample0-prompt_Day_and_night_fight_for_the_domination_of_time}} &
\includegraphics[width=\imwidth]{\impath{batched_samples-style_surrealism-run0-sample3-prompt_Day_and_night_fight_for_the_domination_of_time}} &
\includegraphics[width=\imwidth]{\impath{batched_samples-style_ukiyo_e-run0-sample2-prompt_Day_and_night_fight_for_the_domination_of_time}} \\[-3pt]

\includegraphics[width=\imwidth]{\impath{batched_samples-style_art_nouveau-run0-sample3-prompt_Day_and_night_fight_for_the_domination_of_time}} &
\includegraphics[width=\imwidth]{\impath{batched_samples-style_baroque-run0-sample3-prompt_Day_and_night_fight_for_the_domination_of_time}} &
\includegraphics[width=\imwidth]{\impath{batched_samples-style_expressionism-run0-sample3-prompt_Day_and_night_fight_for_the_domination_of_time}} &
\includegraphics[width=\imwidth]{\impath{batched_samples-style_impressionism-run0-sample2-prompt_Day_and_night_fight_for_the_domination_of_time}} &
\includegraphics[width=\imwidth]{\impath{batched_samples-style_post_impressionism-run0-sample3-prompt_Day_and_night_fight_for_the_domination_of_time}} &
\includegraphics[width=\imwidth]{\impath{batched_samples-style_realism-run0-sample3-prompt_Day_and_night_fight_for_the_domination_of_time}} &
\includegraphics[width=\imwidth]{\impath{batched_samples-style_renaissance-run0-sample2-prompt_Day_and_night_fight_for_the_domination_of_time}} &
\includegraphics[width=\imwidth]{\impath{batched_samples-style_romanticism-run0-sample3-prompt_Day_and_night_fight_for_the_domination_of_time}} &
\includegraphics[width=\imwidth]{\impath{batched_samples-style_surrealism-run0-sample1-prompt_Day_and_night_fight_for_the_domination_of_time}} &
\includegraphics[width=\imwidth]{\impath{batched_samples-style_ukiyo_e-run0-sample3-prompt_Day_and_night_fight_for_the_domination_of_time}} \\
\bottomrule
\end{tabular}
\vspace{-1em}
\caption{\label{fig:finegrained_vis} Visual Examples of fine-grained zero-shot text based stylization with LAION model. The prompt used to generate these samples is \emph{'Day and night fighting for the domination of time'}. More samples provided in the Appendix~\ref{suppsec:add_samples}.\vspace{-2em}}
\end{figure}
}

\newcommand{\classifiereval}{
\vspace{-2em}
\renewcommand{\impath}[1]{img/##1}
\begin{figure}[htbp]
\begin{subfigure}[t]{.5\textwidth}%
\centering
\includegraphics[width=\linewidth]{\impath{percent_improvement}}
\caption{Relative improvement of our proposed approach over postfix-based stylization.}
\label{fig:class_improvement}
\end{subfigure}
\begin{subfigure}[t]{.5\textwidth}%
\centering
\includegraphics[width=\linewidth]{\impath{classifier_voting}}
\caption{Classifier voting results: Our method outperforms the postfix approach for nearly every style.}
\label{fig:class_voting}
\end{subfigure}
\vspace{-0.5em}
\caption{\label{fig:classifier_results} Quantitative comparison of our retrieval and postfix-based prompt stylization. \vspace{-3.25em}}
\end{figure}
}

\newcommand{\cmark}{\ding{51}}%

\newcommand{\hyperparams}{
\begin{table}[htbp]
\centering
\setlength{\tabcolsep}{5pt}
\begin{adjustbox}{max width=\linewidth}
\footnotesize
\begin{tabular}{lcc}
\toprule
& \diffusionmodel & \diffusionmodel   \\
\midrule
Train Dataset & ImageNet  & LAION-2B-en (100M examples)$^\dagger$ \\
image size & $256^2$ & $256^2$ / $512^2$ / $768^2$  \\
$z$-shape (smallest image size) & $64 \times 64 \times 3$ & $16 \times 16 \times 16$ \\
$\vert \mathcal{Z} \vert$ & 8192 & KL  \\
Diffusion steps &1000 & 1000 \\
Noise Schedule & linear&linear\\
Model Size &400M & 1.3B\\
Channels & 192 & 448 \\
Depth &2& 2 \\
Channel Multiplier & 1,2,3,5 & 1,2,3,4 \\
Number of Channels per Head & 32 & 32 \\
Batch Size & 1240 & 1600 / 800 / 432 \\
Iterations & 112K & 275K($256^2$) / 225K($512^2$) / 270K($768^2$) \\
Learning Rate& $\text{1.0e-4}$ &$\text{1.0e-4}$   \\
\midrule
Conditioning & CA & CA   \\
CA-resolutions & 32, 16, 8 & 16, 8, 4  \\
$\phi_{\text{CLIP}}$ model spec & ViT-B/32 & ViT-L/14\\
Embedding Dimension &  512&768\\
Transformer Depth & 1 & 1 \\
\midrule
size $\mathcal{D}_{\text{\tiny train}}$ & 20M & 1.9B \\
$k$ & 4 & 8 \\
$\mathcal{D}_{\text{\tiny train}} \cap \text{Train Dataset} = \emptyset$ & \cmark & \cmark \\ 
image size $\mathcal{D}_{\text{\tiny train}}$ & $256 \times 256$ & $224 \times 224$ \\
\bottomrule
\end{tabular}
\end{adjustbox}
\vspace{1em}
\caption{\label{tab:hyperparams} Hyperparameters for the models introduced in Sec.~\ref{subsec:generalexp}. $^\dagger$: Image size has been successively increased during training, see image sizes}
\end{table}
}

\newcommand{\sampling}{\xi_k}

\begin{document}
\pagestyle{headings}
\mainmatter
\def\ECCVSubNumber{11}  %

%
%
%

%
\title{Text-Guided Synthesis of Artistic Images with Retrieval-Augmented Diffusion Models \vspace*{-0.4cm}} %

\author{Robin Rombach$^*$ \and
Andreas Blattmann$^*$ \and
Björn Ommer}
\authorrunning{R.Rombach \& A.Blattmann et al}
\titlerunning{Denoising in Style} 
\institute{Ludwig-Maximilian University Munich, Germany \\ $^*$\small{\emph{the first two authors contributed equally to this work}}}

%
\maketitle
\enlargethispage{2\baselineskip}
\neatsamples
\newpage
\begin{abstract}
Novel architectures have recently improved generative image synthesis leading to excellent visual quality in various tasks. 
Of particular note is the field of ``AI-Art'', which has seen unprecedented growth with the emergence of powerful multimodal models such as CLIP. By combining speech and image synthesis models, so-called "prompt-engineering" has become established, in which carefully selected and composed sentences are used to achieve a certain visual style in the synthesized image.
In this note, we present an alternative approach based on retrieval-augmented diffusion models (RDMs). In RDMs, a set of nearest neighbors is retrieved from an external database during training for each training instance, and the diffusion model is conditioned on these informative samples. During inference (sampling), we replace the retrieval database with a more specialized database that contains, for example, only images of a particular visual style. This provides a novel way to ``prompt'' a general trained model after training and thereby specify a particular visual style. As shown by our experiments, this approach is superior to specifying the visual style within the text prompt.
%
We open-source code and model weights at \url{https://github.com/CompVis/latent-diffusion}.

\vspace{-1.5em}
\end{abstract}

\noindent Diffusion models have recently set the state of the art in image generation and controllable synthesis~\cite{adm,ldm}. In text-to-image synthesis in particular, we have seen impressive results~\cite{dalle2,imagen} that can also be used to create artistic images. Such models thus have the potential to help artists create new content and have contributed to the tremendous growth of the field of AI generated art~\cite{vqganclip}. However, these models are very compute intensive and so far cannot be reused for tasks other than those for which they were trained.
For this reason, in the present work we build on the recently introduced retrieval-augmented diffusion models (RDMs)~\cite{rdm,knn-diff}, which also have the potential to significantly reduce the computational complexity required in training by providing a comparatively small generative model with a large image database: While the retrieval approach provides the (local) content, the model can now focus on learning the composition of scenes based on this content.   
In this extended abstract, we scale RDMs and show their capability to generate artistic images as those shown in Fig.~\ref{fig:bigsamples}.
Moreover, we can control the synthesis process with natural language by using the joint text-image representation space of CLIP~\cite{clip} and demonstrate that we obtain fine-grained control over the visual style of the output by retrieving neighbors from highly specialized databases built from WikiArt~\cite{wikiart} and ArtBench~\cite{artbench}. Finally, we also consider the release of our model weights as a contribution that allows artists to complement, extend, and evaluate their work and also to investigate the inherent biases of these models.
\vspace{-0.5em}
\section{Recap on Retrieval-Augmented Diffusion Models}
\label{sec:method}
\vspace{-0.5em}
Following \cite{rdm,knn-diff}, a retrieval-augmented diffusion model (\diffusionmodel) is a combination of a conditional latent diffusion model $\epsilon_{\theta}$~\cite{ddpm,ldm}, a database of images $\mathcal{D}_{\text{\tiny train}}$, which is considered to be an explicit part of the model, and a (non-trainable) sampling strategy $\sampling$ to obtain a subset of $\mathcal{D}_{\text{\tiny train}}$ based on a query $x$ as introduced in \cite{rdm}. The model is trained by implementing $\sampling$ as a nearest neighbor algorithm, such that for each query (i.e., training example) its $k$ nearest neighbors are returned as a set, where the distance is measured in CLIP~\cite{clip} image embedding space. 
The CLIP embeddings of these nearest neighbors are then fed to the model via the cross attention mechanism~\cite{transformers,ldm}. The training objective reads
\begin{small}
\begin{equation}
\label{eq:loss}
\min_{\theta} \mathcal{L} = \mathbb{E}_{p(x), z \sim E(x), \epsilon \sim \mathcal{N}(0,1), t}\Big[ \Vert \epsilon - \epsilon_{\theta}\big(z_t, t, \{\phi_{\text{CLIP}}(y) \vert y \in \sampling(x, \mathcal{D}_{\text{\tiny train}})\}\big) \Vert_2^{2} \Big] \, ,
\end{equation}
\end{small}
where $\phi_\text{CLIP}$ is the CLIP image encoder and $E(x)$ is the encoder of an autoencoding model as deployed in~\cite{ldm,rdm}.
After training, we replace $\mathcal{D}_{\text{\tiny train}}$ of the original \diffusionmodel with alternate databases $\mathcal{D}_{\text{\tiny style}}$, derived from art datasets~\cite{wikiart,artbench} to obtain a post-hoc model modification and thereby zero-shot stylization. Furthermore, we can guide the synthesis process with text-prompts by using the shared text-image feature space of CLIP~\cite{clip} as proposed in~\cite{rdm}. Thus, we obtain a \emph{controllable} synthesis model which is only trained on image data.
\vspace{-0.75em}
\section{Text-Guided Synthesis of Artistic Images with RDMs}
\label{sec:exps}
\vspace{-1em}
\subsection{General Setting}
\label{subsec:generalexp}
\vspace{-0.75em}
We conduct experiments for two models: To show the general zero-shot stylization potential of \diffusionmodel, we train an exact replica of the \diffusionmodel on ImageNet~\cite{imagenet} as proposed in~\cite{rdm}, i.e., we build $\mathcal{D}_{\text{\tiny train}}$ from OpenImages~\cite{DBLP:journals/corr/abs-1811-00982}. For inference, we achieve stylization by using a database $\mathcal{D}_{\text{\tiny style}}$ based on the WikiArt~\cite{wikiart} dataset \cf~Sec.~\ref{subsec:stylization}. In Sec.~\ref{subsec:fine_grained} we present a larger model, trained on 100M examples from LAION-2B-en~\cite{schuhmann2021laion400m,laion5bprelim} with a more diverse database $\mathcal{D}_{\text{\tiny train}}$, which contains the remaining 1.9B samples from that dataset. Samples from this model are shown in Fig.~\ref{fig:bigsamples}. By exchanging this database with distinct, style-specific subsets of the ArtBench dataset~\cite{artbench} during inference, we show that \diffusionmodel can further be used for fine-grained stylization, without being trained for this task. 
Details on training and inference are provided in Appendix~\ref{suppsec:model_details}.
\vspace{-1em}
\subsection{Zero-Shot Text-Guided Stylization by Exchanging the Database}
\label{subsec:stylization}
\vspace{-0.75em}
\stylizer
By replacing the train database $\mathcal{D}_{\text{\tiny train}}$ with WikiArt \cite{wikiart} we show the zero-shot stylization capabilities of the ImageNet-\diffusionmodel from in Sec~\ref{subsec:generalexp} in Fig.~\ref{fig:stylize}.
Our model, though only trained on ImageNet, generalizes to this new database and is capable of generating artwork-like images %
which depict the content defined by the text prompts. 
To further emphasize the effects of this post-hoc exchange of the database, we show samples obtained with the same procedure but using $\mathcal{D{\text{\tiny train}}}$ (bottom row). 

\subsubsection{Fine-Grained Stylization with ArtBench}
\label{subsec:fine_grained}
\enlargethispage{1\baselineskip}
Many powerful models emulate text-driven stylization by adding the postfix \emph{''... in the style of ...''} to a given prompt~\cite{ldm,glide,dalle2,imagen,parti}. By using style specific databases obtained from the ArtBench dataset~\cite{artbench} during inference, we here present an alternative approach. 
Fig.~\ref{fig:finegrained_vis} presents results for the prompt \emph{''Day and night fighting for the domination of time.''} and the LAION-\emph{RDM}. Each column contains samples generated by replacing $\mathcal{D{\text{\tiny train}}}$ with a style-specific ArtBench-subset. 
\finegrainedvisual
For a quantitative evaluation, we generate 70 samples per style for both approaches, which we then classify with a style-classifier (details in Appendix~\ref{suppsec:class_details}) and compare both the relative improvement in accuracy (Fig.~\ref{fig:class_improvement}) and the classifier logits (Fig.~\ref{fig:class_voting}). The retrieval-based approach almost always outperforms postfix-based stylization.
\classifiereval
\vspace{-1.5em}
\section{Conclusion}
\vspace{-1.25em}
In this note, we present an approach to train accessible and controllable models for visual art: By building on the recently introduced retrieval-augmented diffusion models, our approach becomes \emph{accessible} as we efficiently store an image database and condition a comparatively small generative model directly on meaningful samples from the database, rather than compressing large training data into increasingly large generative models. Our approach is \emph{controllable} because it allows post-hoc replacement of the external database and thus specification of a desired visual style, which emerges in our experiments as a strong alternative to pure text-based approaches.
In future work, we plan to combine this approach with post-hoc finetuning on paired text-image data.%

\clearpage

\bibliographystyle{splncs04}
\bibliography{ms}

\begin{thebibliography}{10}
\providecommand{\url}[1]{\texttt{#1}}
\providecommand{\urlprefix}{URL }
\providecommand{\doi}[1]{https://doi.org/#1}

\bibitem{laion5bprelim}
Laion-5b: A new era of open large-scale multi-modal datasets.
  \url{https://laion.ai/blog/laion-5b/}, accessed: 2022-07-07

\bibitem{knn-diff}
Ashual, O., Sheynin, S., Polyak, A., Singer, U., Gafni, O., Nachmani, E.,
  Taigman, Y.: Knn-diffusion: Image generation via large-scale retrieval. arXiv
  preprint arXiv:2204.02849  (2022)

\bibitem{rdm}
Blattmann, A., Rombach, R., Oktay, K., Ommer, B.: Retrieval-augmented diffusion
  models. arXiv  (2022)

\bibitem{retro}
Borgeaud, S., Mensch, A., Hoffmann, J., Cai, T., Rutherford, E., Millican, K.,
  Driessche, G.v.d., Lespiau, J.B., Damoc, B., Clark, A., et~al.: Improving
  language models by retrieving from trillions of tokens. arXiv preprint
  arXiv:2112.04426  (2021)

\bibitem{gpt3}
Brown, T., Mann, B., Ryder, N., Subbiah, M., Kaplan, J.D., Dhariwal, P.,
  Neelakantan, A., Shyam, P., Sastry, G., Askell, A., et~al.: Language models
  are few-shot learners. Advances in neural information processing systems
  \textbf{33},  1877--1901 (2020)

\bibitem{ic-gan}
Casanova, A., Careil, M., Verbeek, J., Drozdzal, M., Romero~Soriano, A.:
  Instance-conditioned gan. Advances in Neural Information Processing Systems
  \textbf{34} (2021)

\bibitem{vqganclip}
Crowson, K., Biderman, S., Kornis, D., Stander, D., Hallahan, E., Castricato,
  L., Raff, E.: Vqgan-clip: Open domain image generation and editing with
  natural language guidance (2022). \doi{10.48550/ARXIV.2204.08583},
  \url{https://arxiv.org/abs/2204.08583}

\bibitem{imagenet}
Deng, J., Dong, W., Socher, R., Li, L.J., Li, K., Fei-Fei, L.: Imagenet: A
  large-scale hierarchical image database. In: 2009 IEEE conference on computer
  vision and pattern recognition. pp. 248--255. Ieee (2009)

\bibitem{adm}
Dhariwal, P., Nichol, A.: Diffusion models beat gans on image synthesis.
  Advances in Neural Information Processing Systems  \textbf{34} (2021)

\bibitem{pmlr-v119-guo20h}
Guo, R., Sun, P., Lindgren, E., Geng, Q., Simcha, D., Chern, F., Kumar, S.:
  Accelerating large-scale inference with anisotropic vector quantization. In:
  III, H.D., Singh, A. (eds.) Proceedings of the 37th International Conference
  on Machine Learning. Proceedings of Machine Learning Research, vol.~119, pp.
  3887--3896. PMLR (13--18 Jul 2020),
  \url{https://proceedings.mlr.press/v119/guo20h.html}

\bibitem{guu2020retrieval}
Guu, K., Lee, K., Tung, Z., Pasupat, P., Chang, M.: Retrieval augmented
  language model pre-training. In: International Conference on Machine
  Learning. pp. 3929--3938. PMLR (2020)

\bibitem{ddpm}
Ho, J., Jain, A., Abbeel, P.: Denoising diffusion probabilistic models.
  Advances in Neural Information Processing Systems  \textbf{33},  6840--6851
  (2020)

\bibitem{khandelwal2020nearest}
Khandelwal, U., Fan, A., Jurafsky, D., Zettlemoyer, L., Lewis, M.: Nearest
  neighbor machine translation. arXiv preprint arXiv:2010.00710  (2020)

\bibitem{khandelwal2019generalization}
Khandelwal, U., Levy, O., Jurafsky, D., Zettlemoyer, L., Lewis, M.:
  Generalization through memorization: Nearest neighbor language models. arXiv
  preprint arXiv:1911.00172  (2019)

\bibitem{DBLP:journals/corr/abs-1811-00982}
Kuznetsova, A., Rom, H., Alldrin, N., Uijlings, J.R.R., Krasin, I.,
  Pont{-}Tuset, J., Kamali, S., Popov, S., Malloci, M., Duerig, T., Ferrari,
  V.: The open images dataset {V4:} unified image classification, object
  detection, and visual relationship detection at scale. CoRR
  \textbf{abs/1811.00982} (2018), \url{http://arxiv.org/abs/1811.00982}

\bibitem{artbench}
Liao, P., Li, X., Liu, X., Keutzer, K.: The artbench dataset: Benchmarking
  generative models with artworks. arXiv  (2022)

\bibitem{long2022retrieval}
Long, A., Yin, W., Ajanthan, T., Nguyen, V., Purkait, P., Garg, R., Blair, A.,
  Shen, C., Hengel, A.v.d.: Retrieval augmented classification for long-tail
  visual recognition. arXiv preprint arXiv:2202.11233  (2022)

\bibitem{meng2021gnn}
Meng, Y., Zong, S., Li, X., Sun, X., Zhang, T., Wu, F., Li, J.: Gnn-lm:
  Language modeling based on global contexts via gnn. arXiv preprint
  arXiv:2110.08743  (2021)

\bibitem{glide}
Nichol, A., Dhariwal, P., Ramesh, A., Shyam, P., Mishkin, P., McGrew, B.,
  Sutskever, I., Chen, M.: Glide: Towards photorealistic image generation and
  editing with text-guided diffusion models. arXiv preprint arXiv:2112.10741
  (2021)

\bibitem{clip}
Radford, A., Kim, J.W., Hallacy, C., Ramesh, A., Goh, G., Agarwal, S., Sastry,
  G., Askell, A., Mishkin, P., Clark, J., et~al.: Learning transferable visual
  models from natural language supervision. In: International Conference on
  Machine Learning. pp. 8748--8763. PMLR (2021)

\bibitem{dalle2}
Ramesh, A., Dhariwal, P., Nichol, A., Chu, C., Chen, M.: Hierarchical
  text-conditional image generation with clip latents. arXiv preprint
  arXiv:2204.06125  (2022)

\bibitem{ldm}
Rombach, R., Blattmann, A., Lorenz, D., Esser, P., Ommer, B.: High-resolution
  image synthesis with latent diffusion models. arXiv preprint arXiv:2112.10752
   (2021)

\bibitem{imagen}
Saharia, C., Chan, W., Saxena, S., Li, L., Whang, J., Denton, E., Ghasemipour,
  S.K.S., Ayan, B.K., Mahdavi, S.S., Lopes, R.G., Salimans, T., Ho, J., Fleet,
  D.J., Norouzi, M.: Photorealistic text-to-image diffusion models with deep
  language understanding (2022). \doi{10.48550/ARXIV.2205.11487},
  \url{https://arxiv.org/abs/2205.11487}

\bibitem{wikiart}
Saleh, B., Elgammal, A.M.: Large-scale classification of fine-art paintings:
  Learning the right metric on the right feature. CoRR  \textbf{abs/1505.00855}
  (2015), \url{http://arxiv.org/abs/1505.00855}

\bibitem{schuhmann2021laion400m}
Schuhmann, C., Vencu, R., Beaumont, R., Kaczmarczyk, R., Mullis, C., Katta, A.,
  Coombes, T., Jitsev, J., Komatsuzaki, A.: Laion-400m: Open dataset of
  clip-filtered 400 million image-text pairs (2021)

\bibitem{siddiqui2021retrievalfuse}
Siddiqui, Y., Thies, J., Ma, F., Shan, Q., Nie{\ss}ner, M., Dai, A.:
  Retrievalfuse: Neural 3d scene reconstruction with a database. In:
  Proceedings of the IEEE/CVF International Conference on Computer Vision. pp.
  12568--12577 (2021)

\bibitem{tseng2020retrievegan}
Tseng, H.Y., Lee, H.Y., Jiang, L., Yang, M.H., Yang, W.: Retrievegan: Image
  synthesis via differentiable patch retrieval. In: European Conference on
  Computer Vision. pp. 242--257. Springer (2020)

\bibitem{transformers}
Vaswani, A., Shazeer, N., Parmar, N., Uszkoreit, J., Jones, L., Gomez, A.N.,
  Kaiser, {\L}., Polosukhin, I.: Attention is all you need. Advances in neural
  information processing systems  \textbf{30} (2017)

\bibitem{xu2021texture}
Xu, R., Guo, M., Wang, J., Li, X., Zhou, B., Loy, C.C.: Texture
  memory-augmented deep patch-based image inpainting. IEEE Transactions on
  Image Processing  \textbf{30},  9112--9124 (2021)

\bibitem{parti}
Yu, J., Xu, Y., Koh, J.Y., Luong, T., Baid, G., Wang, Z., Vasudevan, V., Ku,
  A., Yang, Y., Ayan, B.K., Hutchinson, B., Han, W., Parekh, Z., Li, X., Zhang,
  H., Baldridge, J., Wu, Y.: Scaling autoregressive models for content-rich
  text-to-image generation (2022). \doi{10.48550/ARXIV.2206.10789},
  \url{https://arxiv.org/abs/2206.10789}

\end{thebibliography}

\newpage

\FloatBarrier

\clearpage
\newpage

\appendix
\begin{center}
\Huge\textbf{Appendix}
\end{center}

\neatersamples

\section{Related Work}
\label{suppsec:related_work}
\textbf{Retrieval-Augmented Generative Models.} Using external memory to augment traditional models has recently drawn attention in natural language processing (NLP)~\cite{khandelwal2019generalization,khandelwal2020nearest,meng2021gnn,guu2020retrieval}.
For example, RETRO~\cite{retro} proposes a retrieval-enhanced transformer for language modeling which performs on par with state-of-the-art models~\cite{gpt3} using significantly less parameters and compute resources.
These retrieval-augmented models with external memory turn purely parametric deep learning models into semi-parametric ones.
Early attempts~\cite{long2022retrieval,siddiqui2021retrievalfuse,tseng2020retrievegan,xu2021texture} in retrieval-augmented visual models do not use an external memory and exploit the training data itself for retrieval.
In image synthesis, IC-GAN~\cite{ic-gan} utilizes the neighborhood of training images to train a GAN and generates samples by conditioning on single instances from the training data. However, using training data itself for retrieval potentially limits the generalization capacity, and thus, we favor an external memory in this work. For diffusion models, RDM~\cite{rdm} and kNN-diffusion~\cite{knn-diff} pioneer retrieval augmentation.

\section{Details on Presented Models}
\label{suppsec:model_details}
\hyperparams
\subsection{ImageNet Model}
\label{suppsubsec:imagenet_model}
The hyperparameters used for the ImageNet model are presented in Sec.~\ref{subsec:generalexp} and evaluated in Sec.~\ref{subsec:stylization} are presented in the first column of Tab.~\ref{tab:hyperparams}. We here note again that this model in an exact replica of the ImageNet model presented in~\cite{rdm}. 
During training, we use a database $\mathcal{D}_{\text{\tiny train}}$ of 20M examples from the OpenImages~\cite{DBLP:journals/corr/abs-1811-00982} dataset. We extract 2-3 patches per image (more details, see~\cite{rdm})
to use each image at least once. To obtain nearest neighbors we apply the ScaNN search algorithm~\cite{pmlr-v119-guo20h} in the feature space of a pretrained CLIP-ViT-B/32~\cite{clip}. Using this setting, retrieving 20 nearest neighbors from the database described above takes approximately 0.95 ms. The model is trained on eight NVIDIA A-100-SXM4 with 80GB RAM per GPU.

\subsection{LAION Model}
\label{suppsubsec:laion_model}
The hyperparameters used for our LAION model are subsumed in the second column of Tab.~\ref{tab:hyperparams}. For this model, we scale the size of the training dataset as well as those of the database $\mathcal{D}_{\text{\tiny train}}$. The train dataset is obtained by only using images of the LAION-2B-en dataset~\cite{schuhmann2021laion400m,laion5bprelim}, with a shorter edge length larger than $768$ px and filtering images containing watermarks and unsafe content. The database $\mathcal{D}_{\text{\tiny train}}$ is constructed based on the remaining 1.9B images. Hence the database $\mathcal{D}_{\text{\tiny train}}$ is disjoint from the training set. In contrast to the ImageNet model we use the CLIP-ViT-L/14 model both as a retrieval distance and as nearest neighbor encoder $\phi_{\text{CLIP}}$. We train the model in three stages and successively increase the image size of the train images from $256^2$ px in the first to $512^2$ in the second stage before we reach our final resolution of $768^2$ px in the final third stage. The model is trained on eight NVIDIA A-100-SXM4 with 80GB RAM per GPU.

\subsection{Retrieval Strategy During Inference}
\label{suppsubsec:inference_retrieval}
To obtain the nearest neighbors for text-based stylization from the inference-time database $\mathcal{D}_{\tiny style}$ we embed the query prompt into the shared CLIP text-image space by using the CLIP text-encoder~\cite{clip}. We then retrieve the $k=19$ nearest neighbors from $\mathcal{D}_{\tiny style}$. As a distance measure, we use cosine similarity. 

\section{Details on Style-Classifier}
\label{suppsec:class_details}
The style classifier is trained on ArtBench and implemented as a two-layer perceptron on top of CLIP image features. Its top-1 accuracy on the validation set is $77\%$.

\section{Addtional Qualitative Results}
\label{suppsec:add_samples}
\finegrainedsupp
In Fig.~\ref{fig:bigsamples2} we show additional samples from our LAION-model and in Fig.~\ref{fig:finegrained_vis_supp} we provide additional examples showing the style-specific stylization capabilities of this model.
\end{document}